# CENTROG FEATURE TECHNIQUE FOR VEHICLE TYPE RECOGNITION AT DAY AND NIGHT TIMES

Martins E. Irhebhude, Philip O. Odion and Darius T. Chinyio

Faculty of Science, Department of Computer Science, Nigerian Defence Academy, Kaduna, Nigeria.

## ABSTRACT

*This work proposes a feature-based technique to recognize vehicle types within day and night times. Support vector machine (SVM) classifier is applied on image histogram and CENsus Transformed histogRam Oriented Gradient (CENTROG) features in order to classify vehicle types during the day and night. Thermal images were used for the night time experiments. Although thermal images suffer from low image resolution, lack of colour and poor texture information, they offer the advantage of being unaffected by high intensity light sources such as vehicle headlights which tend to render normal images unsuitable for night time image capturing and subsequent analysis. Since contour is useful in shape based categorisation and the most distinctive feature within thermal images, CENTROG is used to capture this feature information and is used within the experiments. The experimental results so obtained were compared with those obtained by employing the CENsus TRansformed hISTogram (CENTRIST). Experimental results revealed that CENTROG offers better recognition accuracies for both day and night times vehicle types recognition.*

## 1. INTRODUCTION

A feature technique that can adequately recognise vehicle types at day and night times has been a subject in most vision related researches. Most of the time, a feature technique that gave an optimal recognition accuracy on day time experiment for vehicle type recognition often fails to replicate the same accuracy result when the same feature is applied on same dataset in same locality at night time. This is probably because, features that were used for the day time experiment are mostly appearance related i.e. information about colour and texture appearance of the vehicle; whereas a night time feature is mostly contour based.

As reported by [1], not much work have been done on vehicle recognition at night time on thermal images. However, the authors proposed CENsus Transformed histogRam Oriented Gradient (CENTROG) feature technique for vehicle recognition at night times. Also reported in [2], Iwasaki et. al., a vehicle detection mechanism within thermal images using the Viola Jones detector was proposed. The technique involved detecting the thermal energy reflection area of tires as a feature. This paper therefore aims to contribute towards bridging this research gap by proposing a technique that will recognise vehicles at both day and night times.

Also reported by [3] and documented by several other authors, vehicle at day time were detected and recognised by using different approaches. A multiple feature techniques was proposed in [3] for recognising vehicle types during the day time. The proposed feature was experimented on two different view angles. A state-of-the-art video processing techniques for vehicle detection, recognition and tracking was surveyed in [4]. A combination of salient geographical and shape features of taillights and license plates extracted from the rear view of a vehicle was used in [5] for the recognition of vehicle make and model. Extracting additional data from video stream, besides the vehicle image itself helped improve vehicle recognition in a 3D image sets [6]. A technique for road intersection classification was proposed in [7]. A comparative analysis between Support Vector Machine (SVM) and deep neural network with sift features showed that automatic feature extraction recorded higher accuracy compared with manual technique [8]. Li *et al* in [9] showed that even in a congested road traffic condition an AND-OR graph (AOG) using bottom-up inference can be used to represent and detect vehicle objects based on both front and rear views. Similarly, [10] proposed the use of strong shadows as a feature to detect the presence of vehicles in a congested environment. In a separate experiment, authors in [11] showed how vehicles were partitioned into three parts; road, head and body, using a tripwire technique. Subsequently Haar wavelet features extracted from each part with PCA performed on features calculated to form 3 category PCA-subspaces. Further, Multiple Discriminate Analysis (MDA) is performed on each PCA-subspace to extract features, which are subsequently trained to identify vehicles using the Hidden Markov Model Expectation Maximisation (HMMEM) algorithm. In an experiment by [12], a camera calibration tool was used on detected and track vehicle objects so as to extract object parameters, which were then used for the classification of the vehicle into classes of cars and non-cars. Vehicle objects were detected and counted using a frame differencing technique with morphological operators: dilation and erosion [13]. In [14], the author added simulated images to initial dataset and applied PCA on each sub-region to reduce feature sets, computation time and hence speed-up the processing cycle. In another classification task in [15], the foreground image size feature was extracted with two levels dilation and fill morphological operations; and classified into small, medium and large categories. In [16], the author proposed Scale-Invariant Feature Transform (SIFT), the Canny edge detector, k-means clustering with Euclidean matching distance metric for inter and intra class vehicle classification as an alternative to expensive Electronic Toll Collection (ETC) full-scale multi-lane free flow traffic system. In [17], a technique for traffic estimation and vehicle classification using region features with a neural network (NN) classifier was proposed. A technique for rear view based vehicle classification was proposed in [18] with investigation of Hybrid Dynamic Bayesian Network (HDBN) in vehicle classification. HDBN was compared with three other classifiers for the classification of known vs unknown classes and four known classes of vehicles using tail light and vehicle dimensions with respect to the dimensions of the license plate as feature sets. Similarly, the width, distance from license plate and the angle between the tail light and the license plates formed part of the eleven features sets.

As we can observe from literature, it is clear that different feature approach has been used for recognising vehicle types at both day and night times respectively. Therefore, in this research work, we will examine features that are contour and appearance based, so as to design a technique that can recognise vehicle during the day and also at night times. This will help eliminate the bottleneck associated with generating new feature sets for dataset at a given time including view angle. Day time video dataset which contain vehicles of varied colours and shapes will be used for the day time experiment while thermal video dataset will be explored for the night time experiment.

The rest of the paper is organized as follows. Section 2 explains the foreground/background segmentation algorithm, Gaussian Mixture Model (GMM), CENsus TRansformed hISTogram (CENTRIST) and CENTROG descriptors along with their usage was discussed in section 3. Section 4 provides an overview of the Support Vector Machine (SVM) classification algorithm. An insight into the research approach is provided in section 5 and followed by experimental and performance evaluation in Section 6. The paper is concluded in Section 7.

## 2. GAUSSIAN MIXTURE MODEL (GMM)

According to [19, 20], a GMM is a parametric probability density function that is represented as a weighted sum of Gaussian distributions. The GMM technique uses a method to model each background pixel by a mixture of $k$ Gaussian distributions [21]. The weight of the mixture represents the time proportion for which the pixel values stay unchanged in a scene. Probable background colours stay longer and are more static than the foreground colours.

In [22], the recent history of each pixel, $X_1,...,X_t$, is modelled by a mixture of $K$ Gaussian distributions. The probability of observing the current pixel value is defined as:

$$P(X_t) = \sum_{i=1}^{K} \omega_{i,t} * \eta(X_t, \mu_{i,t}, \sum_{i,t}) \qquad (1)$$

Where $K$ is the number of distributions, $\omega_{i,t}$ is an estimate of the weight (what portion of the data is accounted for by this Gaussian) of the $i^{th}$ Gaussian in the mixture at time $t$, $\mu_{i,t}$ is the mean value of the $i^{th}$ Gaussian in the mixture at time $t$, $\sum_{i,t}$ is the covariance matrix of the $i^{th}$ Gaussian in the mixture at time $t$, and $\eta$ is a Gaussian probability density function of the form:

$$\eta(X_t, \mu, \sum) = \frac{1}{(2\pi)^{\frac{n}{2}} |\sum|^{\frac{1}{2}}} e^{-\frac{1}{2}(X_t - \mu_t)^T \Sigma^{-1}(X_t - \mu_t)} \qquad (2)$$

The covariance matrix is of the form:

$$\sum_{k,t} = \sigma^2_k I \qquad (3)$$

## 3. CENTRIST AND CENTROG DESCRIPTORS

As reported in [1] Census Transformed Histogram for Encoding Sign Information (CENTRIST) is a visual description technique that was proposed by Wu et. al. [23] that is used to detect topological sections or scene categories. It extracts the structural properties from within an image while filtering out the textural details. It employs the Census Transform (CT) technique in which an 8-bit value is computed in order to encode the signs of comparison between neighbouring pixels. According to [24], CT is a non-parametric local transforms described as follows:

Let $P$ be a pixel, $I(P)$ its intensity (usually an 8-bit integer), and $N(P)$ the set of pixels in some square neighbourhood of diameter $d$ surrounding $P$. All non-parametric transforms depend upon the comparative intensities of $P$ versus the pixels in the neighbourhood $N(P)$.

Define $\xi(P, P')$ to be 1 if $I(P') < I(P)$ and 0 otherwise.

$R\tau(P)$ maps the local neighbourhood surrounding a pixel $P$ to a bit representing the set of neighbouring pixels whose intensity is less than that of $P$. Therefore, census transform compares the intensity value of a pixel with its eight surrounding neighbours; in other words, CT is a summary of local spatial structure given by equation (5) [24]:

Let $N(P) = P$, where $\otimes$ is the Minkowski sum and D is a set of displacements, and let $\otimes$ be concatenation.

$$R\tau(P) = \overset{\otimes}{\underset{[i,\, j]\,\in\, D}{}} \xi(P, P + [i, j]) \qquad\qquad (4)$$

Example:

$$
\begin{array}{|c|c|c|}
\hline
26 & 75 & 65 \\
\hline
26 & 46 & 22 \\
\hline
26 & 40 & 65 \\
\hline
\end{array}
\Rightarrow
\begin{array}{ccc}
1 & 0 & 0 \\
1 & & 1 \\
1 & 1 & 0 \\
\end{array}
\Rightarrow (10011110)_2 \Rightarrow CT = 158
$$

$$
\begin{array}{ccc}
1 & 0 & 0 \\
\Rightarrow 1 & & 1 \\
1 & 1 & 0 \\
\end{array}
\Rightarrow (10011110)_2 \Rightarrow CT = 158 \qquad\qquad (5)
$$

From the CT example above, it can be seen that if the pixel under consideration is larger than (or equal to) one of its eight neighbours, a bit 1 is set in the corresponding location; else a bit 0 is set. The eight bits generated from the intensity comparisons can be put together in the order of appearance (from top to bottom, left to right) and converted to a base-10 value (e.g., binary to decimal conversion). This is the computed CT value for the pixel under consideration. The so-called CENTRIST descriptor therefore is the histogram of the CT image generated from an image.

In [1], in order to compute the CENTROG features, after the image structure has been captured, compute its CT on a computed edge image, thereafter Histogram Oriented Gradient (HOG) [25] features is extracted from the transformed edge image. The HOG works by counting the occurrences of gradient orientation in localized portions of an image. The HOG captures local object appearances and shapes, which can often be characterized rather well by the distribution of local intensity gradients, or edge directions as reported in [26]. The gradient is computed by applying $[1,0,1]$ and $[1,0,1]^T$ in horizontal and vertical directions within an image. The gradient information is collected from local cells and put into histograms using tri-linear interpolation. On the overlapping blocks composed of neighbouring cells, normalisation is performed. The CENTROG descriptor therefore is the HOG on the CT generated image. Some vehicle image samples are shown below, (figures 1 and 2). CENTROG is a very useful technique which helps to capture the local and global structure of a particular image effectively when the colour and texture information are missing in it.

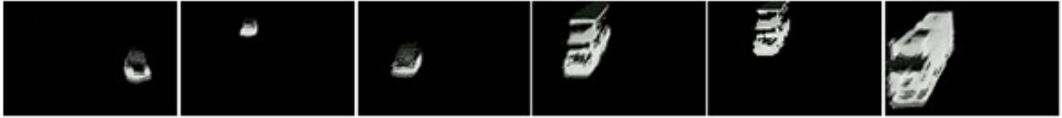

Figure 1: Samples Showing Night Time Vehicles

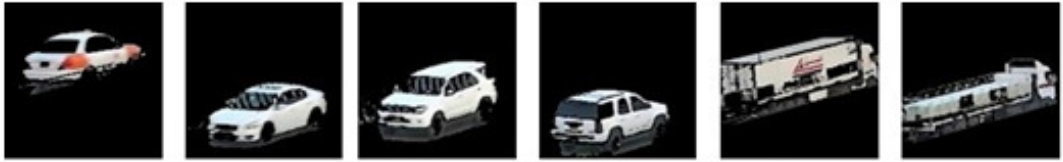

Figure 2: Samples Showing Day Time Vehicles

# 4. SUPPORT VECTOR MACHINE (SVM)

According to [27] SVM is a technique used to train classifiers, regressors and probability densities that is well-founded in statistical learning theory. SVM can be used for binary and multi-classification tasks.

## 4.1. BINARY CLASSIFICATION

SVM perform pattern recognition for two-class problems by determining the separating hyperplane with maximum distance to the closest points of the training set. In this approach, optimal classification of a separable two-class problem is achieved by maximising the width of the margin between the two classes [28]. The margin is the distance between the discrimination hyper-surface in n-dimensional feature space and the closest training patterns called support vectors. If the data is not linearly separable in the input space, a non-linear transformation $\Phi(.)$ can be applied, which maps the data points $x \in R$ into a high dimensional space $H$, which is called a feature space. The data is then separated as described above. The original support vector machine classifier was designed for linear separation of two classes; however, to solve the problem of separating more than two classes, the multi-class support vector machine was developed.

## 4.2 MULTI-CLASS CLASSIFICATION

SVM was designed to solve binary classification problems. In real world classification problems however, we can have more than two classes. In the attempt to solve *q-class* problems with SVMs; training $q$ SVMs was involved, each of which separates a single class from all remaining classes, or training $q^2$ machines, each of which separates a pair of classes. Multiclass classification allows non-linearly separable classes by combining multiple $2-class$ classifiers. $N-class$ classification is accomplished by combining $N$, $2-class$ classifiers, each discriminating between a specific class and the rest of the training set [28]. During the classification stage, a pattern is assigned to the class with the largest positive distance between the classified pattern and the individual separating hyperplane for the N binary classifiers. One of the two classes in such multi-class sets of binary classification problems will contain a substantially smaller number of patterns than the other class [28]. SVM classifier was chosen because of its popularity and speed of processing. For more details on SVM algorithm, please refer to this article [29].

## 5. RESEARCH APPROACH

In order to recognise vehicles at both day and night times, an initial classification between day and night time vehicles will be done. After this initial categorisation, further classification will be done for the day and the night time datasets. Figure 3 shows the proposed methodology for vehicle type recognition at day and night times. As shown in figure 3, the input video was segmented using Gaussian Mixture Model (GMM) foreground/background segmentation algorithm. This was done to enable the effective extraction of the region of interest which contained vehicle objects which were the only moving objects in the video input.

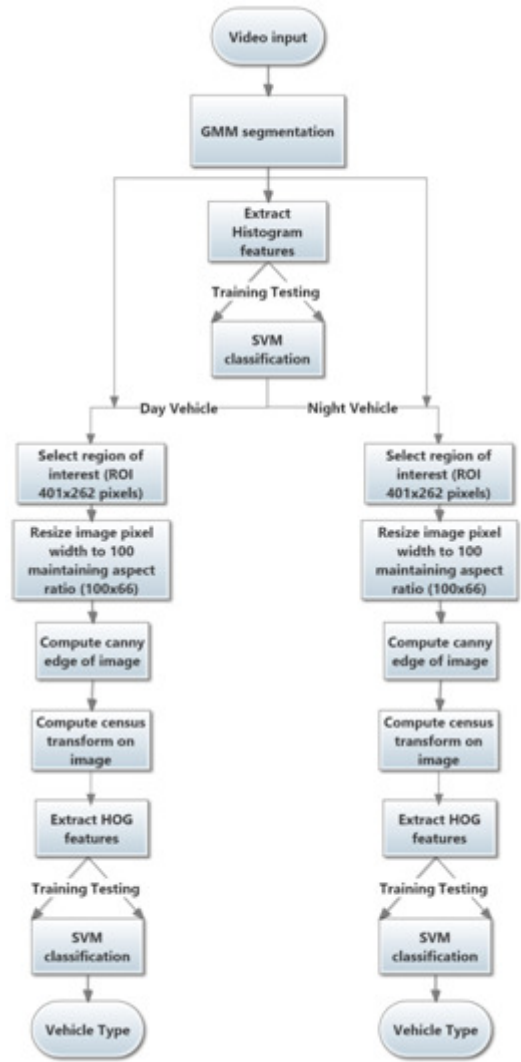

Figure 3: Proposed System Outline for Vehicle Type Recognition at Day and Night Times

In the experiments, based on the number of vehicular samples available in the publicly available night time dataset [30], we will only focus on two key types for night times: cars and trucks. For the night images, we used a thermal camera with parameters FLIR SR-19 Thermal Camera, White Box, Black Box, Total Video Footage Captured: 63 min of ROBB DRIVE and 1-80 OVERPASS. However, for the day vehicular images, data samples were gathered from a low medium resolution camera that was installed on the roadside of the Sohar Highway, in Oman. The camera was of pixel resolutions, 640 x 360, and the frame rate was 25 FPS. The data used in the experimental analysis consisted of 10hours video footage during daytime at approximately a $45^0$ angle from the direction of the movement of vehicles.

## 6. EXPERIMENTS AND PERFORMANCE EVALUATION

A number of experiments were conducted to evaluate the performance of the proposed algorithm on vehicle type recognition at both day and night times. The experiments were conducted on images retrieved from camera installed on a roadside in Sohar Highway, Oman was used for day time vehicular experiments. Similarly, the video dataset given in [30] was used for night time experiments. The results obtain from these experiments will be discussed in the following subsections.

To evaluate the system approach, Receiver Operating Characteristic (ROC) curve will be used. As reported in [3] ROC curve shows classification performance in detail. A ROC curve is the plot of True Positive Rate and the False Positive Rate for different cut-off points or threshold of a parameter [31]. It is given as:

$$\text{True Positive Rate (Recall)} = \frac{tp}{(tp + fn)} \qquad (6)$$

$$\text{False Positive Rate} = \frac{fp}{(fn + tn)} \qquad (7)$$

where, $tp$ denotes the number of true positives (an instance that is positive and classified as positive); $tn$ denotes the number of true negatives (an instance that is negative and classified as negative); $fp$ denotes the number of false positives (an instance that is negative and classified as positive) and $fn$ denotes the number of false negatives (an instance that is positive and classified as negative).

ROC curve visualises the following as reported:

1. It shows the trade off between sensitivity and specificity (any increase in sensitivity will be accompanied by a decrease in specificity).

2. The closer the curve follows the left-hand border and then the top border of the ROC space, the more accurate is the test.

3. The slope of the tangent line at a cutpoint gives the likelihood ratio (LR) for that value of the test.

Similarly, the accuracy of an experiment is measured by the Area Under the ROC Curve (AUC). An area of 1 represents a perfect test; an area of 0.5 or less represents a worthless test. Accuracy of performance is defined as:

$$Accuracy = \frac{t_p + t_n}{t_p + t_n + f_p + f_n} \qquad (8)$$

The following is a rough guide for classifying the accuracy of a test is the traditional academic point system as reported by [32]:

- 0.90-1 = excellent (A)
- 0.80-0.90 = good (B)
- 0.70-0.80 = fair (C)
- 0.60-0.70 = poor (D)
- 0.50-0.60 = fail (F)

Finally, the ROC curve shows the ability of the classifier to rank the positive instances relative to the negative instances.

## 6.1. DAY/NIGHT VEHICULAR CATEGORISATION EXPERIMENTS

The set of input-output sample is

$$(x_1, y_1), (x_2, y_2), ..., (x_N, y_N) \qquad (9)$$

where the input $x_i$ denotes the feature vector extracted from image $i$ and the output $y_i$ is a class label. Since we are categorising into day and night times vehicles, the class label $y_i$ encodes day time vehicles (encoded as 1) and night time vehicles (encoded as 2) respectively; while the extracted feature $x_i$ encodes image histogram. Image histogram feature was used for categorisation because it effectively captures appearance information of any data. Also, there is a clear distinction between day and night data.

The dataset consisted of approximately 1500 day and night time vehicles and was split 75:25 for the purpose of training and testing. The vehicles captured and thus used in experimentation only consisted of cars and trucks (night time vehicles) and cars, jeeps and trucks (day time vehicles) hence the classification was of a binary nature, i.e. into these two classes.

Experiment results shows 100% recognition accuracy; this is due to the fact that there is appearance distinction between the day and night time vehicles. The confusion matrix and ROC curve of the experimental results are shown in figures 4 and 5 below.

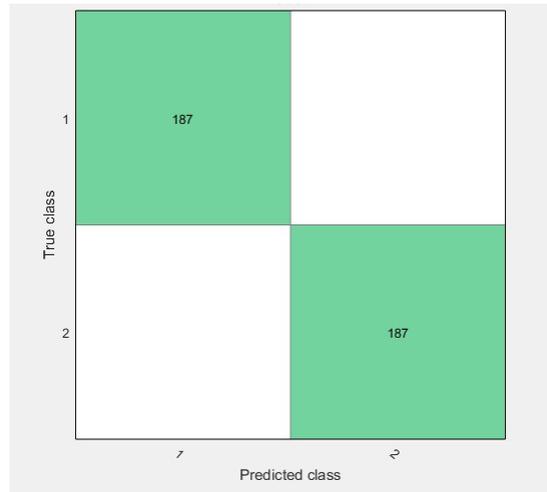

Figure 4: Confusion Matrix for Day and Night Time Categorisation

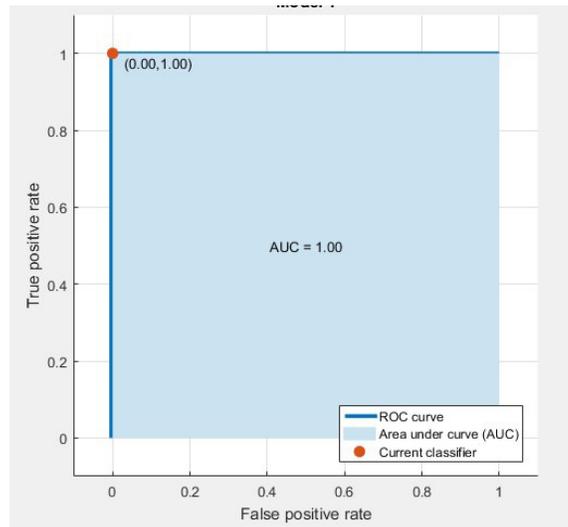

Figure 5: ROC Curve for Day and Night Time Categorisation

## 6.2. Day Vehicular Categorisation Experiments

The set of input-output sample is

$$(x_1, y_1), (x_2, y_2), ..., (x_N, y_N) \qquad (10)$$

where the input $x_i$ denotes the feature vector extracted from image $i$ and the output $y_i$ is a class label. Since we are categorising into vehicle types, the class label $y_i$ encodes cars (encoded as 1), jeeps (encoded as 2), and truck (encoded as 3) respectively; while the extracted feature $x_i$ encodes image CENTROG.

The dataset consisted of approximately 720-day time vehicles and was split 75:25 for the purpose of training and testing. The vehicles captured and thus used in experimentation only consisted of cars, jeeps and trucks hence the classification was of a multi-categorisation, i.e. into three classes.

Experiments were conducted using CENTROG and compared with CENTRIST feature descriptor. Experimental results obtained showed that CENTROG (the proposed technique) outperformed CENTRIST by recording a detection accuracy of 97.2% versus 95.6%. The confusion matrix and ROC curve of the experimental results are shown below in figures 6, 7, 8 and 9 for the purpose of comparison.

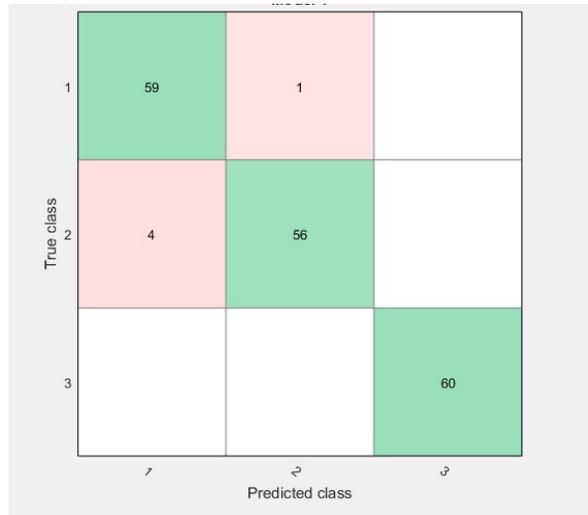

Figure 6: CENTROG Confusion Matrix for Day Time Vehicle Type Recognition

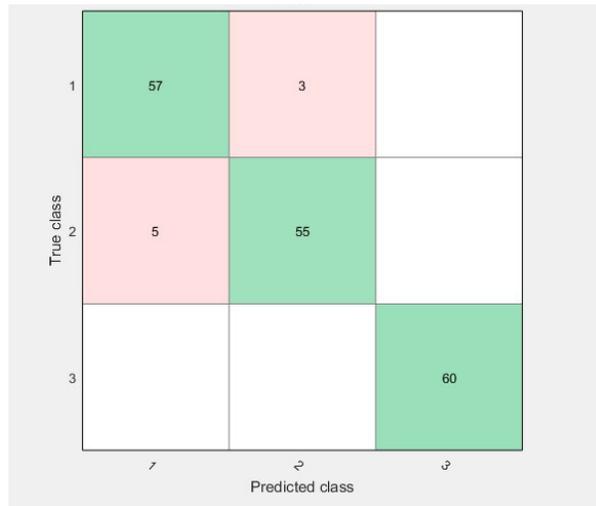

Figure 7: CENTRIST Confusion Matrix for Day Time Vehicle Type Recognition

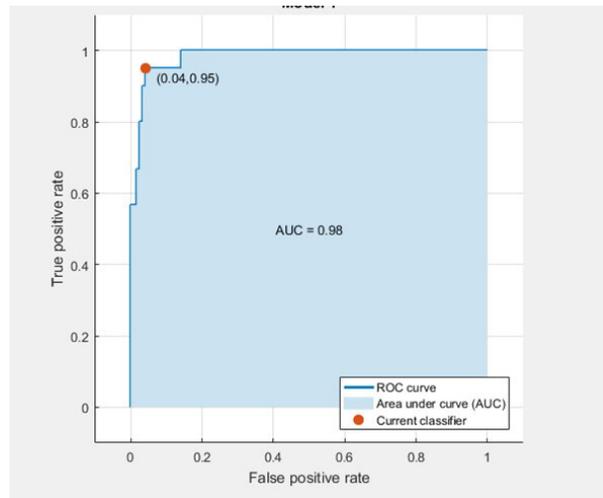

Figure 9: CENTRIST ROC Curve for Day Time Vehicle Type Recognition

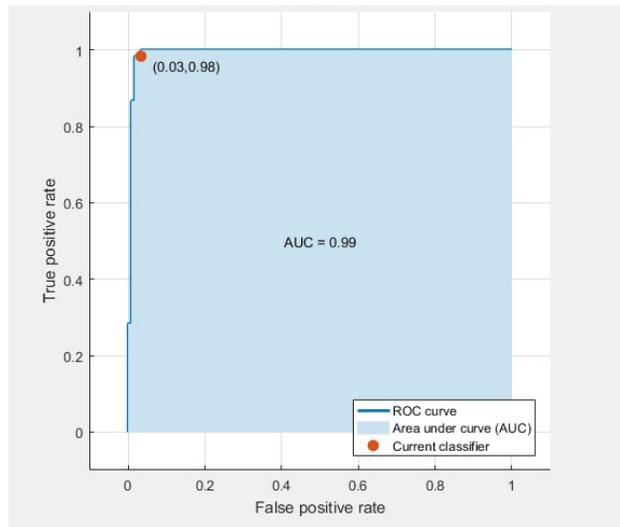

Figure 8: CENTROG ROC Curve for Day Time Vehicle Type Recognition

## 6.3. NIGHT TIME VEHICULAR CATEGORISATION EXPERIMENTS

In the experiments conducted in [1], it was extensively reported that CENTROG feature technique gave highest accuracy results for vehicle type recognition at night time. Results from these experiments showed an accuracy of 100% for the CENTROG technique in contrast to 92.7% for the CENTRIST technique as reported.

## 7. CONCLUSION

In conclusion, this paper proposed a feature-based technique for vehicle type recognition at both day and night times. Initial categorisation was carried out to classify vehicles into day and night time types using image histogram as features. Experiments conducted gave 100% recognition accuracy. In order to recognise the vehicle types the proposed features were extracted by applying Histogram Oriented Gradient on Census Transformed images and hence termed as CENTROG. An SVM classifier was trained on the features obtained from the two datasets (day and night time vehicles). The proposed technique was implemented and compared with the CENTRIST feature technique. Experimental results showed that CENTROG outperformed the CENTRIST, recording 97.2% vs 95.6% (for day time) and 100% vs 92.7% (night time) recognition accuracies respectively, thereby exhibiting a higher classification accuracy.

Future work would involve looking into identifying more categories, such as vans, tricycles and motorcycles etc.

## AUTHORS

**Martins Ekata Irhebhude** obtained his tertiary and master degree education in Edo State, Nigeria in 2003 and 2008 respectively. He is concluded his PhD research degree in 2015 with the Computer Science Department in Loughborough University, UK under the supervision of Eran A. Edirisinghe PhD, a Professor of Digital Image Processing. Martins is a staff of the Nigerian Defence Academy, Kaduna State, Nigeria since 2004 and currently engages in teaching and research activities in and around the Defence Academy. Research interests includes: object detection, people tracking, people re- identification, object recognition and vision related researches.
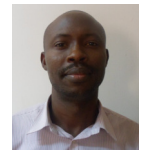

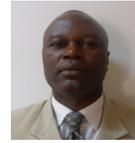

**Philip Oshiokhaimhele Odion** got his first degree (BSc, Computer Science) in 1996 from University of Benin, Benin-city, Edo State-Nigeria. He obtained his MSc, Computer Science at Abubakar Tafawa Balewa University, Bauchi-Nigeria in 2006 and PhD, Computer Science at Nigerian Defence Academy, Kaduna in 2014. Dr PO Odion is currently a Senior Lecturer and Head of Department (HOD) of Computer Science at the Nigerian Defence Academy, Kaduna. He has various local and international publications to his credit. His research interests are in Software Engineering, Computer Networks and Artificial Intelligence.

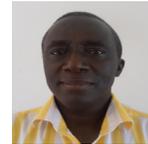

**Darius Tienhua Chinyio** received the B. Ed. degree in Mathematics Education, and a Post Graduate Diploma in Computer Science from Ahmadu Bello University, Zaria, Nigeria, in 1983 and 1986, respectively; and an M. Sc. in Computer Science from the University of Lagos, Nigeria in 1991. He is currently working toward the Ph.D. degree in Computer Science, Department of Computer Science, Nigerian Defence Academy, Kaduna, Nigeria; under the supervision of Professor E. A. Onibere. His research interests include Computational Science, Image Processing, and Networking.